\documentclass[manuscript,screen,nonacm,acmsmall]{acmart}

\AtBeginDocument{%
  \providecommand\BibTeX{{%
    \normalfont B\kern-0.5em{\scshape i\kern-0.25em b}\kern-0.8em\TeX}}}

\usepackage{setspace}

\usepackage{caption} 

\usepackage{float}
\usepackage{graphicx}

\usepackage{tikz}
\usepackage{pgfplots}
\pgfplotsset{compat=newest}
\usepackage{pgfplotstable}

\definecolor{gray1}{HTML}{A4A4A4}

\definecolor{gray4}{HTML}{F2F2F2}

\definecolor{darkgreen}{HTML}{177245}
\definecolor{darkred}{HTML}{990000}

\usepackage{subcaption}
\usepackage{cleveref}
\usepackage{color}

\usepackage{booktabs}
\usepackage{multirow}
\newcommand\mc[1]{\multicolumn{1}{c}{#1}}
\usepackage{array}
\newcolumntype{L}[1]{>{\raggedright\let\newline\\\arraybackslash\hspace{0pt}}p{#1}}
\newcolumntype{C}[1]{>{\centering\let\newline\\\arraybackslash\hspace{0pt}}p{#1}}
\newcolumntype{R}[1]{>{\raggedleft\let\newline\\\arraybackslash\hspace{0pt}}p{#1}}
\newcommand\ie{i.e.}
\newcommand\eg{e.g.}
\usepackage{eurosym}
\usepackage{siunitx}
\usepackage{soul}

\begin{document}

\title{Which linguistic cues make people fall for fake news?\\ A comparison of cognitive and affective processing}

\renewcommand{\shorttitle}{Linguistic cues explain why people fall for fake news}


\author{Bernhard Lutz}
\email{bernhard.lutz@is.uni-freiburg.de}
\affiliation{%
  \institution{University of Freiburg}
  \streetaddress{Rempartstr. 16}
  \city{Freiburg}
  \country{Germany}
}

\author{Marc Adam}
\email{marc.adam@newcastle.edu.au}
\affiliation{%
  \institution{University of Newcastle}
  \streetaddress{Callaghan University Drive}
  \city{Newcastle}
  \country{Australia}
}

\author{Stefan Feuerriegel}
\affiliation{%
  \institution{LMU Munich}
  \streetaddress{Geschwister-Scholl-Platz 1}
  \city{Munich}
  \country{Germany}}
\email{feuerriegel@lmu.de}

\author{Nicolas Pröllochs}
\email{nicolas.proellochs@wi.jlug.de}
\affiliation{%
  \institution{University of Giessen}
  \streetaddress{Licher Straße 74}
  \city{Giessen}
  \country{Germany}
}

\author{Dirk Neumann}
\email{dirk.neumann@is.uni-freiburg.de}
\affiliation{%
  \institution{University of Freiburg}
  \streetaddress{Rempartstr. 16}
  \city{Freiburg}
  \country{Germany}
}

\renewcommand{\shortauthors}{Lutz et al.}

\begin{abstract}
Fake news on social media has large, negative implications for society. However, little is known about what linguistic cues make people fall for fake news and, hence, how to design effective countermeasures for social media. In this study, we seek to understand which linguistic cues make people fall for fake news. Linguistic cues (e.g., adverbs, personal pronouns, positive emotion words, negative emotion words) are important characteristics of any text and also affect how people process real vs. fake news. Specifically, we compare the role of linguistic cues across both cognitive processing (related to careful thinking) and affective processing (related to unconscious automatic evaluations). To this end, we performed a within-subject experiment where we collected neurophysiological measurements of 42 subjects while these read a sample of 40 real and fake news articles. During our experiment, we measured cognitive processing through eye fixations, and affective processing \emph{in situ} through heart rate variability. We find that users engage more in cognitive processing for longer fake news articles, while affective processing is more pronounced for fake news written in analytic words. To the best of our knowledge, this is the first work studying the role of linguistic cues in fake news processing. Altogether, our findings have important implications for designing online platforms that encourage users to engage in careful thinking and thus prevent them from falling for fake news.

\end{abstract}




\maketitle

\newcommand{\highlightcue}[1]{\textsc{#1}}

\section{Introduction}

\raggedbottom


Fake news presents a major threat to social media integrity \citep{Prollochs.2023,Starbird.2019}, with negative implications for the opinions of individuals and society at large \citep{Aral.2019,Bar.2023,Lazer.2018}. As examples, fake news has spread widely during the 2020 US Presidential Election \citep{Flamino.2023,Pennycook.2021b}, the COVID-19 pandemic \citep{Cinelli.2020,Solovev.2022}, and the 2022 Russian invasion of Ukraine \citep{Geissler.2022}. According to global surveys, \SI{86}{\percent} of Internet users have at least once fallen for fake news \citep{CIGI.2019}. To curb the spread, an important prerequisite is to understand why users fall for fake news. 


Prior research has aimed to understand how and why people engage with fake news. For example, research modeled the dynamics behind fake news sharing \citep{Prollochs.2023,Vosoughi.2018,Solovev.2022,EPJ.2021,ScientificReports.2021,Drolsbach.2023,Friggeri.2014}. Other works look at the individual characteristics of fake news recipients and why some people are more likely to believe in fake news. Here, research has looked at different dimensions, such as political orientation \citep{Traberg.2022}, religious animosity \citep{Bronstein.2019}, and trust in science \citep{Scheufele.2019}. Even others look at behavioral aspects such as the extent to which users engage in analytical thinking \citep{Arechar.2022,Pennycook.2021,Pennycook.2019}, as more extensive reasoning should make it easier to debunk false statements. A related stream focuses on how different forms of presentation (e.g., presenting the source of news) affect beliefs \citep[\eg,][]{Spezzano.2021,Gao.2018,Kirchner.2020,Lu.2022}. In contrast to that, we focus on information processing, which is the precursor to belief formation, and thus allows us to examine how users process fake news and why they fall for it.


In this study, we analyze why users fall for fake news based on different linguistic cues. Linguistic cues have been studied extensively in the context of deception detection in online communication \citep[\eg,][]{Conroy.2015,Ho.2015,Siering.2016,Zhou.2004}, providing evidence that deceptive content is qualitatively different from truthful content as deceivers are lacking precise knowledge and details. For instance, deceivers were shown to write messages with lower complexity \citep{Zhou.2004}, but with more words conveying emotions \citep{Ho.2016}. To this end, we posit that linguistic cues trigger different types of information processing, which, in turn, explains why users fall for fake news. 


Here, we compare the role of linguistic cues across both (i)~cognitive processing and (ii)~affective processing of fake news. Cognitive processing refers to careful thinking that users engage in, which is crucial to correctly assess news as real or fake \citep[\eg,][]{Pennycook.2020,Moravec.2022,Arechar.2022}. Affective processing refers to an unconscious automatic evaluation. Specifically, it is a ``neural activity representing the most basic decision-making quality that guides human behavior`` \citep[][p.~148]{Walla.2018}. As such, affective processing can guide human behavior involuntarily and even without users noticing. Of note, affective processing occurs \emph{in situ}, that is, at the moment when users process the content, because of which affective processes should be measured \emph{in situ}, that is, through neurophysiological measurements such as heart rate variability \citep{Appelhans.2006, Lane.2009}. Importantly, prior research has previously shown that the processing of fake news is governed by \emph{both} cognitive processes and affective processes \citep{Lutz.2020,Lutz.2023}, yet it remains unclear how these processes are triggered. Identifying the driving factors of cognitive and affective processing would hence allow for a deeper understanding of why users fall for fake news and may enable the development of more effective mitigation strategies. Our research question is thus: 

\vspace{0.2cm}
\begin{quote}\begin{quote}  
\textsc{Research Question:} \emph{What is the role of linguistic cues (e.g., \highlightcue{adverbs, personal pronouns, positive emotion words, negative emotion words}) in users' (i)~cognitive processing and (ii)~affective processing of fake news?}
\end{quote}\end{quote}
\vspace{0.2cm}


To study the above research question, we conducted a within-subject experiment with $N=42$ subjects that were presented with a sample of 40 news articles of different veracity (\ie, 20 real and 20 fake news articles). During our experiment, we collected neurophysiological measurements. Specifically, we measured (i)~cognitive processing through eye fixations and (ii)~affective processing through heart rate variability. We then perform a regression analysis, where 
we aim to explain cognitive and affective information processing based on different linguistic cues. 
For this purpose, we compare a comprehensive panel of linguistic cues (e.g., \highlightcue{adverbs, personal pronouns, positive emotion words, negative emotion words}). For better readability, we highlight linguistic cues in \highlightcue{small caps}. To the best of our knowledge, this is the first work studying the role of linguistic cues in fake news through a neurophysiological lens.


Our findings reveal that users engage more in cognitive processing of fake news articles when they are longer and written with lower complexity. In contrast, we find that users experience affective processing when fake news articles are written with more words linked to analytic thinking. A possible reason for the latter is that analytic words may make a fake news article appear more truthful and thus trigger an affective state of psychological discomfort. Our findings can assist online platforms in implementing countermeasures against fake news. Specifically, algorithmic curation at social media platforms is desirable that incentivizes cognitive processing, which, in turn, suggests that platforms should particularly prefer elaborate social media posts over short posts and posts with few analytical thinking words. This would promote cognitive thinking among users and make them more resilient against fake news. On the other hand, analytical words are especially dangerous as they may trigger affective processing, because of which users can fall for fake news inadvertently.


Our \textbf{main contributions} are as follows:
\begin{enumerate}
\item We study how linguistic cues in online news articles are linked to cognitive and affective processing of fake news. To the best of our knowledge, this is the first work studying the role of linguistic cues in fake news processing. 
\item We provide results from a within-subject experiment where users process a sample of 20 real and 20 fake news during which we collected neurophysiological measurements. 
\item We discuss important implications of our work for designing online platforms so that they encourage users to engage in careful thinking and thus prevent them from falling for fake news.
\end{enumerate}

\section{Background}

We first provide a background on fake news (\Cref{sec:rw_fake_news}) and human information processing (\Cref{sec:rw_information_processing}). We then review works at the intersection, that is, information processing of real vs. fake news (\Cref{sec:rw_information_processing_fake_news}). To understand differences in the information processing of real vs. fake news, we then motivate our choice of linguistic cues as important determinants (\Cref{sec:rw_linguistic_cues}).

\subsection{Fake News}
\label{sec:rw_fake_news}

Fake news has been defined as news articles that are verifiably false and, could mislead readers \citep{Lazer.2018}. There are several forms of fake news, including, fabricated content about terrorism or natural disasters, alleged claims about products, or false statements discrediting political rivals. As such, fake news provides a considerable threat to how users form opinions and thus to the functioning of our society \citep{Lazer.2018}.

Several works have aimed to understand fake news along different dimensions capturing human judgment and decision-making. The considered dimensions include, among others, how exposure to fake news influences existing beliefs \citep [\eg,][]{Lazer.2019,Pennycook.2018}, which motivational and socio-demographic factors make users fall for fake news \citep[\eg,][]{Pennycook.2020,Pennycook.2018}, and how fake news propagates \citep[\eg,][]{Solovev.2022,Drolsbach.2023,Prollochs.2023,Hopp.2020,Vosoughi.2018}. For detailed literature reviews about research on fake news, we refer to \citet{Aghajari.2023} and \citet{Pennycook.2021}. Our work extends current research on fake news by disentangling the linguistic cues that guide human information processing, as detailed in the following.

\subsection{Information Processing}
\label{sec:rw_information_processing}

Information processing starts after a piece of information is translated into signals that can be processed by the brain \citep{Walla.2018}. One then distinguishes between (1)~cognitive processing and (2)~affective processing. Cognitive processing leads to an understanding of \emph{what} something is, while affective processing is evaluative, leading to a decision on \emph{how} something is \citep{LeDoux.1989}. According to \citet{Walla.2018}, affective processing forms the basis of any human behavior. Therefore, affective processing also has a salient influence on how humans respond to fake news. 
Importantly, cognitive and affective processing occur in different brain regions. Cognitive processing occurs in the cortical brain regions, while affective processing occurs in the subcortical brain regions \citep{Walla.2018}. With language being a cortical brain function, self-reports cannot fully reflect on the processes that occur deep inside the brain but instead must be measured \emph{in situ} through neurophysiological measurements \citep{vomBrocke.2020,Walla.2018}. 


To generate a better understanding of human interaction with technology, research has increasingly made use of neurophysiological data \citep{vomBrocke.2020}. Neurophysiological data from the human body has been repeatedly shown to help explain the variance in human behavior \citep{Zhang.2013}. Employing measures from neuroscience in particular allows researchers to gain a deeper understanding of the underlying cognitive and affective processes in human-computer interaction \citep{Dimoka.2011}. Specifically, neurophysiological measurements address limitations of self-reports as they can capture these processes \emph{in situ}, i.e., directly at the moment they occur.


\textbf{Cognitive processing:} Cognitive processing can be measured with eye tracking \citep{Rayner.1998}. When reading, the eyes quickly move from one position to another. Between these movements, the eyes are relatively stable, which is referred to as a fixation \citep{Rayner.1978}. Fixations denote short pauses of 200--250ms where the textual information is processed \citep{Rayner.1998}. The number of eye fixations hence provides a measure of cognitive processing \citep{Just.1980,Loftus.1972,Rayner.1978}. Hence, we later measure cognitive processing through eye fixations.


\textbf{Affective processing:} Affective processing can be measured in terms of heart rate variability (HRV) \citep{Appelhans.2006,Thayer.2012}. The neuropsychological relationship between affective processing and HRV is as follows. The human body continuously processes external stimuli in order to accommodate the entire system to environmental demands. Thereby, the human body prepares a regulated response to these external stimuli through the autonomic nervous system (ANS). The ANS is regulated by the hypothalamus, which is located in the subcortical brain regions \citep{Uylings.2000} and it is divided into two subsystems, namely, the inhibitory parasympathetic nervous system (``rest and digest'') and an excitatory sympathetic nervous system (``flight or fight'') \citep{Appelhans.2006}. These branches often interact antagonistically to produce varying degrees of physiological arousal. During psychological discomfort, the balance between these branches (\ie, sympathovagal balance) is shifted to a state where the sympathetic branch becomes dominant. This is accompanied by lower HRV \citep{Pumprla.2002}. Hence, we later measure affective processing through HRV.

\subsection{Information Processing of Real and Fake News}
\label{sec:rw_information_processing_fake_news}

Research on information processing of fake news is still in its infancy. Recent work has shown that the processing of fake news is governed by both cognitive processes and affective processes \citep{Lutz.2020,Lutz.2023}. However, it remains unclear how these processes are triggered. Below, we provide an overview of prior research on the role of cognition and affect in the context of fake news.

\textbf{Cognitive processing:} Cognitive processing is considered a necessary requirement for correctly identifying fake news \citep{Moravec.2019,Pennycook.2020b,Pennycook.2020}. If users engage in cognitive processing by thinking critically about the content, they are less likely to fall for fake news. This effect is even independent of personal attitudes and partisanship \citep{Pennycook.2020}. Therefore, researchers try to identify ways to make users engage in cognitive processing when reading fake news. However, the crux of identifying ways of making users think critically about the content is that users generally tend to minimize cognitive effort \citep{Taylor.1978}. The tendency to spend cognitive effort further depends on users' mindset. Users on social media are generally in a hedonic mindset, where they are seeking pleasure and entertainment, rather than spending cognitive effort into separating real from fake news \citep{Moravec.2019}.

\textbf{Affective processing:} As discussed above, affective processing is typically contrasted with cognitive processing as it occurs automatically and unconsciously \citep{vomBrocke.2020,Walla.2018}. It might thus provide a key determinant of why users perceive online news articles as real or fake and, eventually, explain why users fall for fake news. The results of recent studies suggest that affective processing can indeed occur when users are presented with fake news \citep{Lutz.2020,Lutz.2023}, yet it remains unclear why affective processing occurs. Here, we seek to explain the processing of fake news through linguistic cues.

\subsection{Linguistic Cues}
\label{sec:rw_linguistic_cues}

Linguistic cues (e.g., \highlightcue{adverbs, personal pronouns, positive emotion words, negative emotion words}) provide an important tool to infer textual features and thus characterize \emph{how} texts are written. Crucially, linguistic cues distinguish truthful from deceptive content such as fake news \citep{Gravanis.2019}. Authors of fake news, for obvious reasons, do not have factual details about the fake story they are about to tell. Hence, they can provide less precise information about their fake story \citep{Newman.2003,Zhou.2004}. 

The role of linguistic cues has been studied in the context of human behavior, yet outside of fake news. Here, we give several examples to motivate our focus on linguistic cues as determinants of how users process fake news. For instance, in the context of online news consumption, \highlightcue{positive emotion words} (e.g., \emph{love, nice, sweet}) and negative emotion words (e.g., \emph{hurt, ugly, nasty}) increase the consumption of the corresponding online news \citep{NHB.2023}. In general, \highlightcue{emotion words} may elicit cognitive processes such as attention \citep[\eg,][]{Kissler.2007,Smith.1996}. For instance, messages in discussion forums with more \highlightcue{emotion words} receive more feedback than messages with fewer \highlightcue{emotion words} \citep{Huffaker.2010}. Emotion words are also fixated more often than other words during reading \citep{Scott.2012}. In contrast, longer articles appear more persuasive as they provide more arguments in favor of the advocated opinion \citep{Schwenk.1986,Tversky.1974}. Likewise, \highlightcue{analytic cues} (e.g., \emph{think, know, but, else}) may thus have a deterrent effect on users' willingness to engage in cognitive processing. Motivated by this, we later analyze a wide, comprehensive array of different linguistic cues.

\subsection{Research Gap}

Prior research has shown that the processing of fake news is governed by both cognition and affect, yet it has remained unclear how these processes are triggered. Here, we perform a within-subject experiment with \emph{in-situ} neurophysiological measurements to understand how linguistic cues in fake news are linked to cognitive and affective processing. To the best of our knowledge, this is the first work studying the role of linguistic cues in real and fake news through a neurophysiological lens.

\section{Method}

\begin{figure}[htp]
    \centering
    \includegraphics[trim=0.5cm 5.5cm 0.5cm 1cm, width=0.99\textwidth]{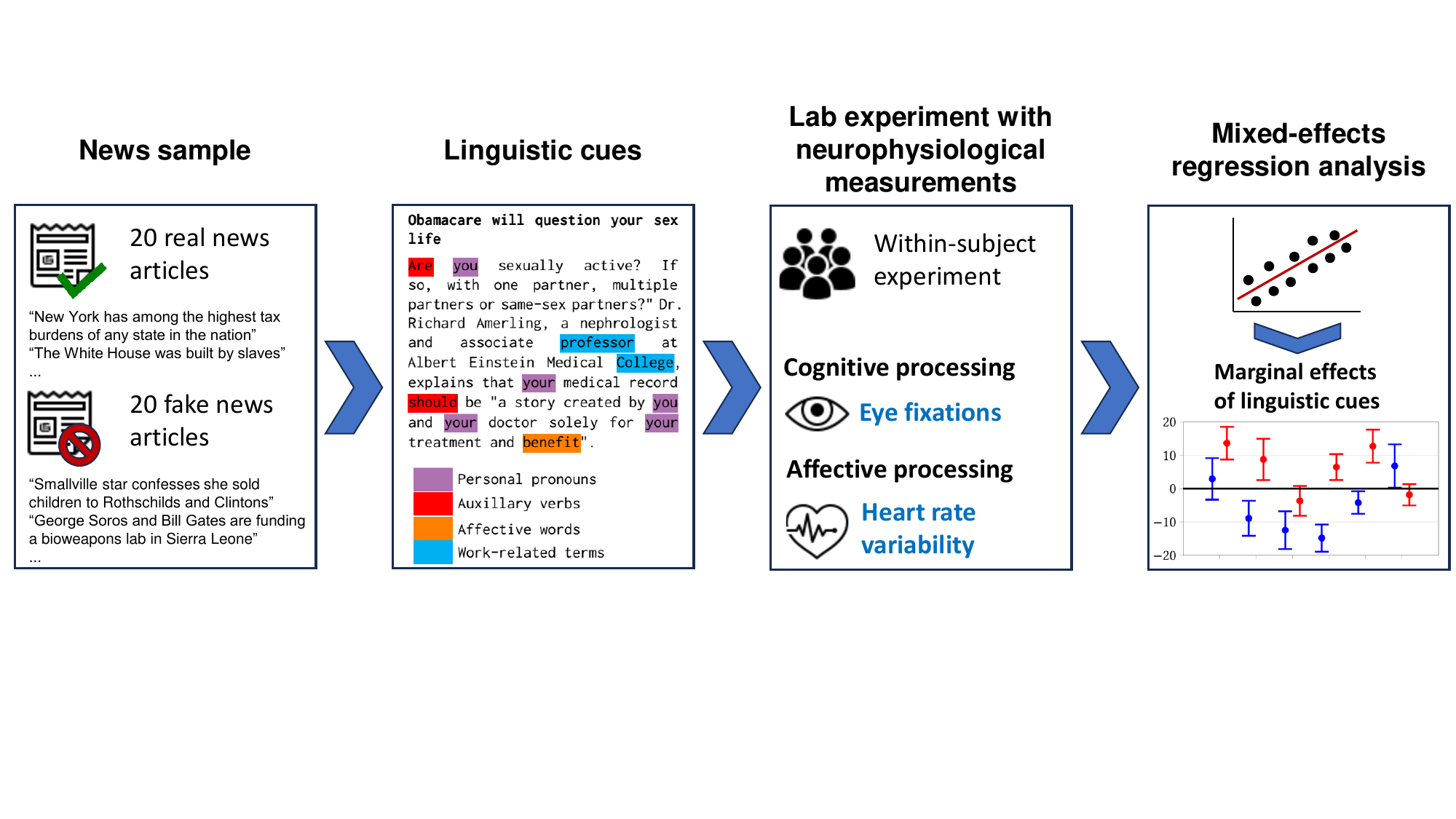}  
    \caption{Research framework.}
    \label{fig:framework}
\end{figure}

The objective of this study is to understand how linguistic cues in real and fake news are linked to the recipients' cognitive and affective processing. The overall research framework is shown in \Cref{fig:framework}. We first collect a sample of 40 real and fake news articles. Second, we calculate a comprehensive set of linguistic cues. Third, we perform a within-subject experiment with neurophysiological measurements.\footnote{Collected data from this experiment was also used for the evaluation of another study \citep{Lutz.2020,Lutz.2020.Cues,Lutz.2023}.} Each subject was shown our sample of real and fake news in random order. During this, we measure (1)~the number of eye fixations via eye tracking as a proxy for cognitive processing, and (2)~heart rate variability via electrocardiography as a proxy for affective processing. Finally, we use the experimental data to estimate mixed-effects regression models to obtain the marginal effects of linguistic cues on cognitive and affective processing.


\subsection{Materials}


Our sample of 20 real and 20 fake news articles originates from the fact-checking website ``Politifact.'' Politifact is a common choice in related studies for retrieving fake news articles and fact-checks \citep{Allcott.2017,Lazer.2018,Vosoughi.2018}. Politifact builds upon trained journalists to investigate political statements and online news articles to assign a veracity label that ranges from ``true'' for real news to ``pants on fire'' for fake news \citep{Graves.2016}. Following \citet{Vosoughi.2018}, we validated the Politifact labels against other fact-checking websites, namely,  snopes.com, factcheck.org, truthorfiction.com, hoax-slayer.com, and urbanlegends.about.com. We found a perfect pairwise agreement (\ie, Fleiss kappa of 1.0) for articles that are also listed on other sites. Details on our news sample, including their veracity labels and URLs to the fact-checking websites are provided in Appendix A of the Supplementary Materials.


In line with previous research \citep[\eg,][]{Bago.2022,Pennycook.2019,Pennycook.2020b}, the  experimental objective is to differentiate between real news articles published by mainstream media and fake news articles disseminated through alternative media outlets. This choice is natural as mainstream media is generally considered unlikely to intentionally create fake news \citep{Tsfati.2020}. Although it remains uncertain whether the alternative media outlets intentionally publish fake news, existing studies suggest that these outlets exhibit a discernible disregard for truth \citep{Grinberg.2019}, aligning with the definition of fake news proposed by \citet{Lazer.2018}.


Note that we do not manipulate the cues manually in our news as this would change the underlying content and would no longer reflect actual news articles as they appear ``in the wild.'' Hence, a strength of our experimental design is to avoid such interventions but, instead, ensure that we use real-world news articles.

\subsection{Calculation and Selection of Linguistic Cues}


We follow prior research \citep[\eg,][]{Ho.2016,Newman.2003} in calculating linguistic cues with ``Linguistic Inquiry and Word Count'' (LIWC) \citep{Pennebaker.2015}. This software is based on a dictionary that contains more than 4,500 words, each of which is classified into one or more dimensions of linguistic cues. A linguistic cue is then calculated as the fraction of words linked to a particular dimension (e.g., \highlightcue{positive emotion words, we-references, certainty}) over all words of a text. For better readability, we highlight linguistic cues in \highlightcue{small caps}.

Given that LIWC contains more than 90 linguistic cues, we focus on an informative subset, which allows us to make meaningful inferences. Specifically, we focus on those cues for our evaluation, for which the average fraction of matched words over all news articles is at least 1\%. In addition, we include the summary variables \highlightcue{tone}, \highlightcue{analytic thinking} \citep{Pennebaker.2014}, and \highlightcue{authentic} \citep{Newman.2003} as these are calculated based on multiple individual cues. Furthermore, we include the \highlightcue{length} (in words) and \highlightcue{complexity} measured as Gunning-Fog readability score \citep{Gunning.1968}. Following this approach, we obtain a total of 33 linguistic cues as shown in \Cref{tbl:cues_description}. The table provides the corresponding LIWC dimension and several examples according to \citet{Pennebaker.2015}.

\begin{table}[htp]
\centering
\caption{Descriptions of linguistic cues (sorted as in the LIWC manual \citep{Pennebaker.2015}).  \label{tbl:cues_description}}
\footnotesize
\setlength{\defaultaddspace}{1.5pt}
\begin{tabular}{>{\scshape}L{3cm}lL{2.5cm}}
\toprule

\textnormal{\textbf{Linguistic cue}} & \textbf{LIWC dimension} & {\textbf{Example words}}  \\

\midrule

Length & \texttt{WC} &  -- \\ \addlinespace 
Complexity & \texttt{Gunning-Fog index$^a$} & -- \\ \addlinespace 

Personal pronouns & \texttt{ppron} & \emph{I, them, her} \\ \addlinespace 
3rd-person singular pronouns & \texttt{shehe} & \emph{she, her, him} \\ \addlinespace 
3rd-person plural pronouns & \texttt{they} & \emph{they, their, they’d} \\ \addlinespace 
Impersonal pronouns & \texttt{ipron} & \emph{it, it’s, those}  \\ \addlinespace 
Auxiliary verbs & \texttt{auxverb} & \emph{am, will, have} \\ \addlinespace 
Adverbs & \texttt{adverb} & \emph{very, really}  \\ \addlinespace 
Comparisons & \texttt{compare} & \emph{greater, best, after} \\ \addlinespace 
Interrogatives & \texttt{interrog} & \emph{how, when, what} \\ \addlinespace 
Numbers & \texttt{number} & \emph{second, thousand}  \\ \addlinespace 
Quantifiers & \texttt{quant} & \emph{few, many, much}  \\ \addlinespace 
Affective processes & \texttt{affect} & \emph{happy, cried}  \\ \addlinespace 
Positive emotion words & \texttt{posemo} & \emph{love, nice, sweet}  \\ \addlinespace 
Negative emotion words & \texttt{negemo} & \emph{hurt, ugly, nasty}  \\ \addlinespace 

Male references & \texttt{male} & \emph{boy, his, dad}  \\ \addlinespace 
Insight & \texttt{insight} & \emph{think, know}  \\ \addlinespace 
Causation & \texttt{cause} & \emph{because, effect} \\ \addlinespace 
Tentative & \texttt{tentat} & \emph{maybe, perhaps}  \\ \addlinespace 
Differentiation & \texttt{differ} & \emph{hasn’t, but, else} \\ \addlinespace 
Perception & \texttt{percept} & \emph{look, heard, feeling}  \\ \addlinespace 
See & \texttt{see} & \emph{view, saw, seen}  \\ \addlinespace 
Biological processes & \texttt{bio} &  \emph{eat, blood, pain} \\ \addlinespace 
Health & \texttt{health} & \emph{clinic, flu, pill} \\ \addlinespace 
Affiliation & \texttt{affiliation} &  \emph{ally, friend, social} \\ \addlinespace 
Achievement & \texttt{achieve} & \emph{win, success, better} \\ \addlinespace 
Power & \texttt{power} & \emph{superior, bully} \\ \addlinespace 
Motion & \texttt{motion} & \emph{arrive, car, go} \\ \addlinespace 
Work & \texttt{work} & \emph{job, majors, xerox} \\ \addlinespace 
Money & \texttt{money} & \emph{audit, cash, owe} \\ \addlinespace 

Analytic thinking & \texttt{Analytic} & -- \\ \addlinespace 
Authenticity & \texttt{Authentic} & -- \\ \addlinespace 
Tone & \texttt{Tone} & -- \\ \addlinespace 

\bottomrule
\multicolumn{3}{l}{Note: Examples are taken from the LIWC 2015 manual  \citep{Pennebaker.2015}.} \\
\multicolumn{3}{l}{$^a$ Gunning-Fox readability index is calculated using the} \\
\multicolumn{3}{l}{``quanteda'' package in R.}\\
\end{tabular}

\end{table}


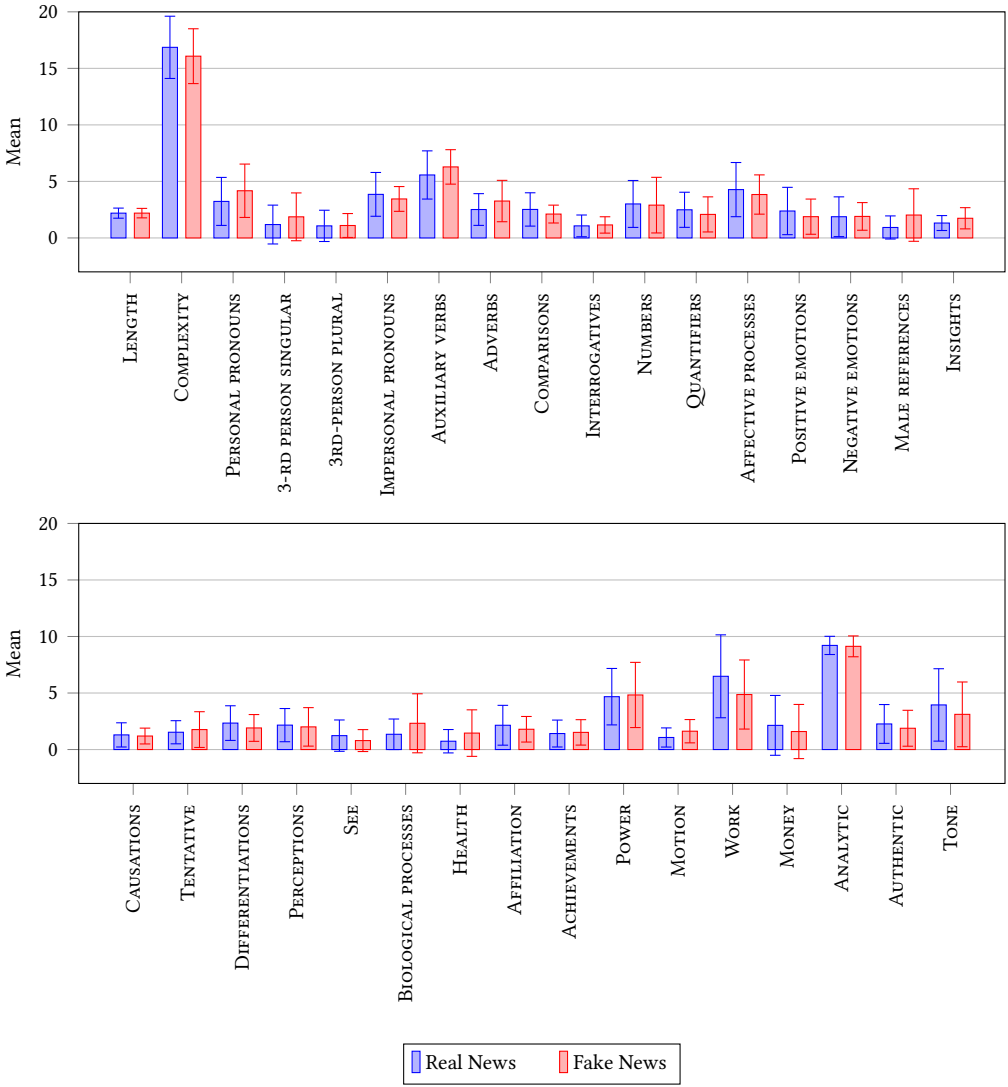
\begin{figure}[htp]
    \centering

    \begin{subfigure}[b]{\textwidth} {
    \begin{tikzpicture}
\pgfplotsset{
  error bars/.cd,
    x dir=none,
    y dir=both, y explicit,
}
\begin{axis}[
    width=\linewidth,
    height=5cm, ybar,
    no marks,
    bar width=0.2cm,
    ymajorgrids,
    tick pos=lower,
    tick align=outside,
    xmin=0, xmax=18,
    ymin=-3, ymax=20,
    ylabel style={font=\footnotesize},
    xlabel style={font=\footnotesize},
    ytick={0,5,10,15,20},
    xtick={1,2,3,4,5,6,7,8,9,10,11,12,13,14,15,16,17},
xticklabels={\highlightcue{Length},\highlightcue{Complexity},\highlightcue{Personal pronouns},\highlightcue{3-rd person singular},\highlightcue{3rd-person plural},\highlightcue{Impersonal pronouns},\highlightcue{Auxiliary verbs},\highlightcue{Adverbs},\highlightcue{Comparisons},\highlightcue{Interrogatives},\highlightcue{Numbers},\highlightcue{Quantifiers},\highlightcue{Affective processes},\highlightcue{Positive emotions},\highlightcue{Negative emotions},\highlightcue{Male references},\highlightcue{Insights}},
    y tick label style={font=\footnotesize},
    x tick label style={rotate=90, font=\footnotesize},
    ylabel={Mean},
    legend style={
		at={(0.5,-1)},
		anchor=north,
  font=\footnotesize,
		legend columns=-1,
		/tikz/every even column/.append style={column sep=0.5cm}
	},
  ]
  \addplot+ [mark=*] table [x expr={\thisrow{x}-0.03}, y=y, y error=ey] {
    x y ey
1 2.193 0.447 0.447
2 16.860 2.753 2.753
3 3.231 2.123 2.123
4 1.185 1.718 1.718
5 1.068 1.383 1.383
6 3.854 1.937 1.937
7 5.571 2.132 2.132
8 2.514 1.404 1.404
9 2.521 1.474 1.474
10 1.068 0.961 0.961
11 3.007 2.075 2.075
12 2.490 1.553 1.553
13 4.276 2.396 2.396
14 2.383 2.093 2.093
15 1.877 1.758 1.758
16 0.926 1.023 1.023
17 1.318 0.660 0.660

  };
  \addplot+ [mark=*] table [x expr={\thisrow{x}+0.03}, y=y, y error=ey] {
    x y ey
1 2.195 0.419 0.419
2 16.078 2.425 2.425
3 4.174 2.360 2.360
4 1.868 2.115 2.115
5 1.100 1.057 1.057
6 3.449 1.096 1.096
7 6.283 1.524 1.524
8 3.264 1.830 1.830
9 2.112 0.790 0.790
10 1.150 0.724 0.724
11 2.902 2.460 2.460
12 2.082 1.552 1.552
13 3.841 1.737 1.737
14 1.882 1.557 1.557
15 1.904 1.223 1.223
16 2.026 2.322 2.322
17 1.740 0.938 0.938

  };
  
\end{axis}
\end{tikzpicture}
}

\end{subfigure}
    \begin{subfigure}[b]{\textwidth} {
\pgfplotsset{compat=1.11,
		/pgfplots/ybar legend/.style={
			/pgfplots/legend image code/.code={%
				\draw[##1,/tikz/.cd,yshift=-0.32em]
				(0cm,0cm) rectangle (3pt,0.8em);},
		},
	}
    \begin{tikzpicture}
\pgfplotsset{
  error bars/.cd,
    x dir=none,
    y dir=both, y explicit,
}
\begin{axis}[
    width=\linewidth,
    height=5cm,ybar,
    no marks,
    bar width=0.2cm,
    ymajorgrids,
    tick pos=lower,
    tick align=outside,
    xmin=0, xmax=17,
    ymin=-3, ymax=20,
    ytick={0,5,10,15,20},
    xtick={1,2,3,4,5,6,7,8,9,10,11,12,13,14,15,16},
xticklabels={\highlightcue{Causations},\highlightcue{Tentative},\highlightcue{Differentiations},\highlightcue{Perceptions},\highlightcue{See},\highlightcue{Biological processes},\highlightcue{Health},\highlightcue{Affiliation},\highlightcue{Achievements},\highlightcue{Power},\highlightcue{Motion},\highlightcue{Work},\highlightcue{Money},\highlightcue{Analytic},\highlightcue{Authentic},\highlightcue{Tone}},
    y tick label style={font=\footnotesize},
    x tick label style={rotate=90, font=\footnotesize},
    xlabel style={font=\footnotesize},
    ylabel={Mean},
    ylabel style={font=\footnotesize},
    legend style={
		at={(0.5,-1)},
		anchor=north,
  font=\footnotesize,
		legend columns=-1,
		/tikz/every even column/.append style={column sep=0.5cm}
	},
  ]
  \addplot+ [mark=*] table [x expr={\thisrow{x}-0.03}, y=y, y error=ey] {
    x y ey
1 1.296 1.070 1.070
2 1.530 1.023 1.023
3 2.340 1.533 1.533
4 2.160 1.468 1.468
5 1.230 1.386 1.386
6 1.350 1.350 1.350
7 0.736 1.034 1.034
8 2.148 1.762 1.762
9 1.416 1.190 1.190
10 4.676 2.496 2.496
11 1.067 0.847 0.847
12 6.479 3.667 3.667
13 2.138 2.647 2.647
14 9.211 0.805 0.805
15 2.264 1.716 1.716
16 3.947 3.202 3.202

  };
  \addplot+ [mark=*] table [x expr={\thisrow{x}+0.03}, y=y, y error=ey] {
    x y ey
1 1.195 0.700 0.700
2 1.766 1.581 1.581
3 1.910 1.185 1.185
4 2.005 1.702 1.702
5 0.793 0.970 0.970
6 2.322 2.610 2.610
7 1.455 2.058 2.058
8 1.794 1.131 1.131
9 1.518 1.124 1.124
10 4.828 2.883 2.883
11 1.622 1.030 1.030
12 4.867 3.056 3.056
13 1.594 2.395 2.395
14 9.129 0.922 0.922
15 1.883 1.589 1.589
16 3.113 2.863 2.863

  };
  
\legend{Real News, Fake News}
\end{axis}
\end{tikzpicture}
}
\end{subfigure}
    \caption{Barplot showing the computed values for the linguistic cues as means. The error bars denote the standard deviation.}
    \label{fig:summary_barplot}
\end{figure}

\subsection{Theoretical Justification}


The selected linguistic cues are associated with different psychological indicators \citep{Tausczik.2010}. For instance, deceivers tend to write their messages with lower \highlightcue{complexity} as it requires cognitive effort to produce deceptive content. Similarly, the usage of \highlightcue{tentative language} (\emph{maybe, perhaps, guess}) signals uncertainty in the sense that a story is not yet fully established. Conversely, the usage of \highlightcue{differentiation} words signals that the author can precisely distinguish one aspect of the story from another \citep{Newman.2003}. Deceivers also tend to use more \highlightcue{negative emotion words}, as well as less third person and \highlightcue{first person singular pronouns} \citep{Bond.2005,Newman.2003,Zhou.2004}. From an author's perspective, \highlightcue{personal pronouns} can be used to guide the reader's attention. \citet{Gunsch.2000} found that positive political ads often refer to the own candidate or party using more self-references (\emph{I, we}), whereas negative political ads (e.g., about an opponent) tend to use more \highlightcue{3rd-person pronouns} (\emph{he, she, they}). Attention can also reflect how the author is processing the described event. \citet{Kowalski.2000} found that students used more \highlightcue{first-person singular} and fewer \highlightcue{3rd-person pronouns} when describing an event when they were being teased compared to describing an event were they were teasing someone else. Finally, the usage of emotion words tells how authors are experiencing the world. \highlightcue{Positive emotion words} (\emph{nice, sweet, cute}) are more frequently used to describe positive events, while \highlightcue{negative emotion words} (\emph{ugly, brutal, horrible}) are more frequently used to describe negative events \citep{Kahn.2007}. The usage of emotion words may hence reflect how the author copes with an event and how much the event will affect the future. Note that these explanations only explain the psychological underpinnings of a subset of our cues selection. For more comprehensive explanations, we refer the reader to \citet{Tausczik.2010}. 

\subsection{Summary Statistics}


\Cref{fig:summary_barplot} shows the computed values for all considered cues for real news (in blue) and fake news (in red). We observe that the linguistic cues of real and fake news in our sample have a similar frequency. Detailed numeric summary statistics are provided in Appendix B of the Supplementary Materials.


We provide an illustrative example of one real and one fake news article in \Cref{fig:example}. We highlighted the top-5 most frequent linguistic cues (\highlightcue{personal pronouns, impersonal pronouns, auxillary verbs, affective words}, and \highlightcue{work-related words}) in different colors. The example also shows that the LIWC 2015 dictionaries used to calculate linguistic cues are not mutually exclusive. Instead, different linguistic cues often share several words. For instance, the word \emph{benefits} occurs in the list of \highlightcue{affective words}, \highlightcue{positive emotion words}, and \highlightcue{work-related terms}.


Therefore, we also consider the correlations of linguistic cues across all news articles. We find that several linguistic cues exhibit high positive correlations. The top-3 correlations across all news articles occur between \highlightcue{positive emotion words} and \highlightcue{tone} ($r=0.800, p < 0.001$), \highlightcue{see} and \highlightcue{perception} ($r=0.796, p < 0.001$), and \highlightcue{biological processes} and \highlightcue{health} ($r=0.774, p < 0.001$). Correlation plots are shown in Appendix~C of the Supplementary Materials. We later account for the correlations of linguistic cues by including only a single linguistic cue in each regression model. By estimating one regression model for each cue, the coefficients are thus not affected by multicollinearity issues.


\definecolor{Orchid}{HTML}{AF72B0}

\newcommand{\myviolet}{Orchid}
\newcommand{\ctextlila}[1]{%
  \begingroup
  \sethlcolor{\myviolet}%
  \hl{#1}%
  \endgroup
}

\newcommand{\mygreen}{green}
\newcommand{\ctextgreen}[1]{%
  \begingroup
  \sethlcolor{\mygreen}%
  \hl{#1}%
  \endgroup
}

\newcommand{\myred}{red}
\newcommand{\ctextred}[1]{%
  \begingroup
  \sethlcolor{\myred}%
  \hl{#1}%
  \endgroup
}
\newcommand{\myorange}{orange}
\newcommand{\ctextorange}[1]{%
  \begingroup
  \sethlcolor{\myorange}%
  \hl{#1}%
  \endgroup
}

\definecolor{mylightblue}{HTML}{00B0F0}
\newcommand{\myblue}{mylightblue}
\newcommand{\ctextblue}[1]{%
  \begingroup
  \sethlcolor{\myblue}%
  \hl{#1}%
  \endgroup
}

\begin{figure}
	\footnotesize
	\ttfamily
 \singlespacing

 \fbox{\begin{subfigure}[t]{0.45\textwidth}
 {\rmfamily\emph{\guillemotright}}\textbf{Is There a Right to Social Security?
} \vspace{0.2cm}\\ 
\ctextlila{You} \ctextblue{worked} hard \ctextlila{your} whole life and paid thousands of dollars in Social Security \ctextblue{taxes}. Now \ctextred{it's} time to \ctextblue{retire}. \ctextlila{You}\ctextred{'re} \ctextblue{llegally} entitled to Social Security benefits, right? Wrong. There \ctextred{is} no \ctextblue{legal} right to Social Security, and \ctextgreen{that} \ctextred{is} one of the considerations \ctextgreen{that} \ctextred{may} decide the coming debate over Social Security reform.  [...] Ephram Nestor \ctextred{was} a Bulgarian immigrant \ctextgreen{who} came to the United States in 1918 and paid Social Security \ctextblue{taxes} from 1936, the year the system began operating, until \ctextlila{he} 
retired in 1955. A year after \ctextlila{he} \ctextblue{retired}, Nestor \ctextred{was} deported for \ctextred{having been} a member of the Communist Party in the 1930s. In 1954 Congress \ctextred{had} passed a law saying \ctextgreen{that} any person deported from the United States \ctextred{should} \ctextorange{lose} \ctextlila{his} Social Security \ctextorange{bene}\ctextblue{fits}. Accordingly, Nestor's \$55.60 per month Social Security checks \ctextred{were} stopped.  
[...]{\rmfamily\emph{\guillemotleft}} \vspace{0.2cm}\\
Source: \url{https://www.cato.org/publications/commentary/is-there-right-social-security}
 \caption{Real news.}
 \end{subfigure}}
 \hspace{0.5cm}
  \fbox{\begin{subfigure}[t]{0.45\textwidth}
   {\rmfamily\emph{\guillemotright}}\textbf{Obamacare will question your sex life} \vspace{0.2cm}\\
\ctextred{Are} \ctextlila{you} sexually active? If so, with one partner, multiple partners or same-sex partners?" \ctextred{Be} ready to answer \ctextgreen{those} questions and more the next time \ctextlila{you} go to the doctor, whether \ctextred{it's} the dermatologist or the cardiologist and no matter if the questions \ctextred{are} unrelated to why \ctextlila{you're} seeking medical help. And \ctextlila{you} \ctextred{can} thank the Obama health law. "This \ctextred{is} \ctextorange{nasty} business," says New York cardiologist Dr. Adam Budzikowski. \ctextlila{He} called the sex questions "insensitive, \ctextorange{stupid} and very intrusive." \ctextlila{He} \ctextred{couldn't} think of an occasion when a cardiologist \ctextred{would} need such information - but \ctextlila{he} knows \ctextlila{he'll} \ctextred{be} pushed to ask for it. [...] Doctors and hospitals \ctextgreen{who} \ctextred{don't} comply with the federal government's electronic-health-records requirements forgo incentive \ctextblue{payments} now; starting in 2015, \ctextlila{they'll} face \ctextblue{financial} penalties from Medicare and Medicaid.  
[...]{\rmfamily\emph{\guillemotleft}} \vspace{0.2cm} \\
Source: \url{https://nypost.com/2013/09/15/obamacare-will-question-your-sex-life/}
 \caption{Fake news.}
 \end{subfigure}}
	\normalfont
\vspace{0.3cm}

 

	\begin{tikzpicture}
	    [every node/.style={draw=none,anchor=north west, minimum height=0.6cm, minimum width=1cm}]
  \matrix [draw=black]
  {
    \node[fill=\myviolet] { \ }; & \node {\highlightcue{Personal pronouns}};\\
    \node[fill=\mygreen] { \ }; & \node {\highlightcue{Impersonal pronouns}};\\
    \node[fill=\myred] { \ }; & \node {\highlightcue{Auxillary verbs}};\\
    \node[fill=\myorange] { \ }; & \node {\highlightcue{Affective words}};\\
    \node[fill=\myblue] { \ }; & \node {\highlightcue{Work-related terms}};\\
  };
	\end{tikzpicture}
 
 \caption{Illustrative examples of real and fake news articles. Linguistic cues are highlighted in different colors.}
	\label{fig:example}
\end{figure}
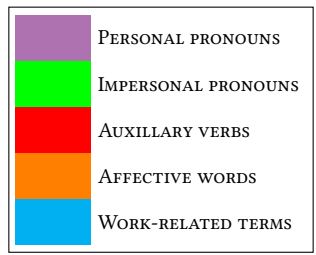

\subsection{Procedure}


The procedure of our experiment was as follows. After signing the consent form, each subject took a comfortable sitting position in our laboratory and was asked to turn off all electronic devices. Subsequently, the electrodes of the electrocardiography (ECG) device were placed on the subject's arms. The calibration and tests of the measurement devices concluded the preparation. The subject was then presented with a dummy news article in order to become familiar with the experimental setting. Afterwards, we scheduled a break of five minutes, such that the subject's HRV normalizes.

The subject was then presented with the aforementioned sample of 40 news articles in random order. The news articles were displayed in a generic format on a 22-inch monitor with 1920x1080 resolution; the distance to the monitor was approx. \SI{60}{cm}. Each news article was presented in two steps. In the first step, only the headline was shown, and, based on this, the subject assessed the veracity solely based on the headline (from 0=strongly real to 6=strongly fake). In the second step, the subject read the news body and categorized the article as either real or fake. This two-step approach was chosen as users generally see the headline first and only subsequently the news body. Subjects earned \euro0.50~(approx.~USD~0.60)~for each correct news categorization; however, they did not receive feedback on whether their answers were correct. Instead, the total amount was only displayed at the end of the experiment. During the experimental procedure, the subject had to keep their arms in a fixed position in order to ensure correct ECG measurements. For the same reason, all decisions needed to be announced to the supervisor, who then entered the answers. For all measurements, the experimental platform ``Brownie'' \citep{Hariharan.2017} was used.


After the experiment, the subject filled out a survey and provided the following information: age, gender, cultural background as European or non-European, disposition to trust based on six questions on 7-point Likert scales from 0=low to 6=high \citep{Gefen.2004}, news consumption from 0=low to 6=high on a 7-point Likert scale, and political attitude towards social and economic issues on 7-point Likert scales from 0=strongly liberal to 6=strongly conservative \citep{Pennycook.2018}. In our analysis, we later account for subject heterogeneity by using a random-effect model, but we use the socio-demographic data as part of our robustness checks.

\subsection{Participants}

The required number of participants was determined using ``G*Power'' \citep{Faul.2009}. Assuming an effect size of $d=0.10$ (due to neurophysiological measurements), $alpha = 0.05$, and $power = 0.80$, the power analysis suggested a minimum number of 36 participants. The participants were recruited during lectures and exercises, as well as, via student mailing lists. Ethics approval was reviewed and granted by the ethics committee at the University of Freiburg.


A total of \textit{N}~$=42$ subjects participated in our study. We had to discard two subjects and an additional two observations from the remaining subjects, where the sensor measurements failed. The final dataset thus consisted of 1598 observations from 40 subjects. All subjects were college students in a first- or second-degree program, 26~male and 14~female with a mean age of \SI{26.0}{years}~(SD\,=\,3.33).

\subsection{Measures}


During the whole experiment, we measured (1)~cognitive processing as the number of eye fixations with the eye tracker ``Tobi Pro X3-120'' and (2)~affective processing as heart rate variability from the ECG measurement device ``BioSignals Plux Solo.'' Both represent later our dependent variables:
\begin{enumerate}
\item \emph{\textbf{Cognitive processing ($\hat{=}$eye fixations)}}. While reading, the number of eye fixations presents a measure for cognitive processing \citep{Rayner.1978,Rayner.1998}. Our eye tracker measured the relative coordinates $(x,y)$ for both eyes on the screen between $0$ and $1$. We removed all observations, where at least one data point was not available, which can happen when users blink or when their gaze is outside the screen. Subsequently, we calculated the number of eye fixations with the package ``saccades'' in R. The package computes the fixations with the velocity-based algorithm from \citep{Engbert.2003}. However, note that an increased number of eye fixations does not necessarily imply a more accurate veracity assessment. Instead, eye fixations measure cognitive effort, which may also increase when users are unsure about whether an article is real or fake.

\item \emph{\textbf{Affective processing ($\hat{=}$heart rate variability)}}. Affective states of discomfort are reflected in lower heart rate variability \citep{Appelhans.2006}. This is because lower HRV reflects dominance of the sympathetic nervous system, which prepares the human body for a ``flight or fight'' response. We calculated HRV as follows. First, we performed an automated detection of the Q, R, and S waves to identify the heartbeats in the ECG signal \citep{Astor.2013}. The heartbeats are denoted by the most characteristic R-peaks. The detection was subsequently validated and corrected during manual inspection with the software ``ECGEditor.'' Second, we used the tool ``cmetx''~\citep{Allen.2007} to calculate HRV as standard deviation in milliseconds of normal-to-normal inter-beat intervals (SDNN) as this is considered the most common HRV metric \citep{Pumprla.2002}. For instance, the average SDNN of resting females between 20 and 26 years is 49.51 milliseconds. During situations of discomfort, SDNN can decrease to 32.70 milliseconds \citep{Pereira.2017}. 

\end{enumerate}

\section{Empirical Analysis}


Our final dataset consists of 1,598 observations from 40 subjects and 20 real and 20 fake news articles (recall that we had to delete two observations during preprocessing). Overall, the subjects identified the veracity correctly in \SI{73.47}{\percent} of all observations. This suggests that separating real from fake news in our sample of news articles was challenging, even for subjects with higher education.

\subsection{Regression Model}


The objective of this study is to explain how linguistic cues are linked to cognitive and affective processing. To this end, we estimate linear mixed-effects models with two dependent variables, namely, subjects' eye fixations (cognitive processing) and HRV (affective processing). More eye fixations indicate that subjects engage more in cognitive processing, while lower HRV indicates that users experience an affective state of discomfort.


The explanatory variables of interest are the linguistic cues. We hence estimate regression models that use linguistic cues to explain eye fixations or HRV. To avoid biased coefficients due to multicollinearity issues, we estimate separate regression models for each cue, each including one of the 33 cues from \Cref{tbl:cues_description}. In each model, we also include an interaction term Cue $\times$ Veracity, which allows us to quantify whether the marginal effects of linguistic on subjects' cognitive/affective processing differs across fake news vs. real news. The dummy variable Veracity equals 1 for fake news, and 0 for real news.


We include the following control variables in each model. First, we include the veracity dummy variable as real and fake news articles may directly be linked to cognitive and affective processing. Second, we control for the sequence number (0--39) that specifies when the news article was shown during the experiment. Including the sequence number into our model allows us to account for effects regarding habituation and fatigue. Third, we control for the veracity assessment of an article that was formed based on the headline (from 0=strongly real to 6=strongly fake). This is based on the idea that users typically see the headline before seeing the body. Accordingly, the assessment towards the headline may influence how the body is processed. Fourth, we control for the time in seconds that subjects required to read the news body. Given that we measure cognitive processing as the number of eye fixations, a longer reading time is likely to yield a greater number of fixations. Fifth, we control for the length and complexity of a news article as longer or harder to read texts may directly influence how information is processed. Complexity is measured as the Gunning-Fog index \citep{Gunning.1968}, which indicates the number of years of formal education required to comprehend the text. Accordingly, greater complexity scores indicate less readable texts. We standardize all explanatory variables in order to assess marginal effects. Accordingly, the coefficients of the linguistic cues denote how the dependent variable changes if the corresponding variable is increased by one standard deviation. 


Finally, we control for subject heterogeneity by including a subject-specific random intercept. The reason is that humans may have general differences in how much they engage in cognitive processing, resulting in additional variance in the number of eye fixations that cannot be explained by other variables. Furthermore, heart rate is regulated differently across different individuals, and some individuals are, in general, more or less likely to experience an affective state of discomfort. Different from fixed effects, including a random effect still allows us to analyze the between-group variation given by how subjects' cognitive and affective processing varies between real and fake news \citep{Townsend.2013}.


Taken together, the resulting linear mixed-effect regression models explaining eye fixations (cognitive processing) and HRV (affective processing) for a particular linguistic cue are given as


\begin{align}
    \text{eye fixations} &= \alpha  + \beta_1 \,\text{Veracity} + \beta_2 \, \text{Sequence number} + \beta_3 \, \text{Headline belief} \nonumber \\ &+\beta_4\, \text{Reading time} + \beta_5 \, \text{Length} + \beta_6 \, \text{Complexity} + \beta_7 \, \text{Length} \times \text{Veracity} \nonumber \\  &+ \beta_8 \, \text{Complexity} \times \text{Veracity}  +  \gamma_1 \, Cue + \gamma_2 \, \text{Cue} \times \text{Veracity} + \delta_U + \varepsilon, \label{eq:model_cog} \\
    \text{HRV} &= \alpha  + \beta_1 \,\text{Veracity} + \beta_2 \, \text{Sequence number} + \beta_3 \, \text{Headline belief} \nonumber \\ &+\beta_4\, \text{Reading time} + \beta_5 \, \text{Length} + \beta_6 \, \text{Complexity} + \beta_7 \, \text{Length} \times \text{Veracity} \nonumber \\  &+ \beta_8 \, \text{Complexity} \times \text{Veracity}  +  \gamma_1 \, Cue + \gamma_2 \, \text{Cue} \times \text{Veracity} + \delta_U + \varepsilon, \label{eq:model_aff}
\end{align}

\vspace{0.3cm}
\noindent
with intercept $\alpha$, subject-specific random intercept $\delta_U$, and error term $\varepsilon$. Hence, we are primarily interested in the coefficients $\gamma_1$ and $\gamma_2$. The marginal effect of linguistic cues for the processing of real news is given as $\gamma_1$, while the marginal effect for this cue for the processing of fake news is given as $\gamma_1 + \gamma_2$.

Importantly, we followed best practices in regression modeling to ensure the validity of our estimates. Among other checks, we calculated the variance inflation factors for all variables to check for possible multicollinearity (and ensured that all values are below the critical threshold of 4). All models are estimated using the ``lmer'' function of the ``lme4'' package in R 4.1.2.

\subsection{Empirical Results}

\subsubsection{Linguistic Cues and Cognitive Processing}

\begin{figure}[htp]
    \centering

    \begin{subfigure}[b]{\textwidth} {

    \begin{tikzpicture}
\pgfplotsset{
  error bars/.cd,
    x dir=none,
    y dir=both, y explicit,
}
\begin{axis}[
    width=\linewidth,
    height=5cm,
    only marks,
    ymajorgrids,
    tick pos=lower,
    tick align=outside,
    xmin=0, xmax=18,
    ymin=-20, ymax=20,
    xtick={1,2,3,4,5,6,7,8,9,10,11,12,13,14,15,16,17},
xticklabels={\highlightcue{Length},\highlightcue{Complexity},\highlightcue{Personal pronouns},\highlightcue{3-rd person singular},\highlightcue{3rd-person plural},\highlightcue{Impersonal pronouns},\highlightcue{Auxiliary verbs},\highlightcue{Adverbs},\highlightcue{Comparisons},\highlightcue{Interrogatives},\highlightcue{Numbers},\highlightcue{Quantifiers},\highlightcue{Affective processes},\highlightcue{Positive emotions},\highlightcue{Negative emotions},\highlightcue{Male references},\highlightcue{Insights}},
    y tick label style={font=\footnotesize},
    x tick label style={rotate=90, font=\footnotesize\scshape},
    xlabel style={font=\footnotesize},
    ylabel style={text width=4cm,align=center,font=\footnotesize},
    ylabel={Marginal effect on cognitive processing},
    legend style={
		at={(0.5,-1)},
		anchor=north,
  font=\footnotesize,
		legend columns=-1,
		/tikz/every even column/.append style={column sep=0.5cm}
	},
  ]
  \addplot+ [mark=*] table [x expr={\thisrow{x}-0.2}, y=y, y error=ey] {
    x y ey
1 -3.55366320479017 4.23719400049046
2 6.61542187305609 4.00260379397233
3 -1.75998414142847 4.28019125790288
4 -0.261055095691601 4.44789693618725
5 -4.21630993789738 3.39560603589646
6 -2.4061982271689 3.478194319423
7 -1.56546298693503 4.61465644743997
8 -7.41526501795621 4.66703714375905
9 3.90760703457762 3.23852567180735
10 -4.16524773441034 3.67022711333196
11 -3.85588270917335 4.40257749296518
12 1.52535741604856 4.17372778238333
13 8.51217865196381 3.91181870048255
14 4.11622821632898 3.72251096779205
15 4.82376641731791 3.46239365774235
16 9.51837884816939 7.46653402567946
17 -0.435313359963968 5.04013188792025

  };
  \addplot+ [mark=*] table [x expr={\thisrow{x}+0.2}, y=y, y error=ey] {
    x y ey
1 3.1020169855351 4.01497731794744
2 -6.32794725553105 4.1180183672506
3 -1.79638078487394 3.97241166903632
4 -0.587359923384806 3.70988128184291
5 -3.12560000871205 4.47120211680722
6 4.03211339041398 5.53193192010245
7 -2.76751166399269 4.83327686986433
8 -3.36540998054951 3.48473526741334
9 4.42924084780877 5.91840866050598
10 -3.2071420172741 5.04485269511156
11 -2.57151860805654 3.58006691882555
12 -1.63247000794627 4.12134831509007
13 -2.05409857247557 5.65716169014618
14 -1.93147947937579 5.24987752051545
15 -0.298062436692157 4.8163633912691
16 -2.75881056199456 3.62005668489901
17 -1.4722486150987 3.44181310667103

  };
  
\end{axis}
\end{tikzpicture}
}

\end{subfigure}
    \begin{subfigure}[b]{\textwidth} {

    \begin{tikzpicture}
\pgfplotsset{
  error bars/.cd,
    x dir=none,
    y dir=both, y explicit,
}
\begin{axis}[
    width=\linewidth,
    height=5cm,
    only marks,
    ymajorgrids,
    tick pos=lower,
    tick align=outside,
    xmin=0, xmax=17,
    ymin=-20, ymax=20,
    xtick={1,2,3,4,5,6,7,8,9,10,11,12,13,14,15,16},
xticklabels={\highlightcue{Causations},\highlightcue{Tentative},\highlightcue{Differentiations},\highlightcue{Perceptions},\highlightcue{See},\highlightcue{Biological processes},\highlightcue{Health},\highlightcue{Affiliation},\highlightcue{Achievements},\highlightcue{Power},\highlightcue{Motion},\highlightcue{Work},\highlightcue{Money},\highlightcue{Analytic},\highlightcue{Authentic},\highlightcue{Tone}},
    y tick label style={font=\footnotesize},
    x tick label style={rotate=90, font=\footnotesize},
    xlabel style={font=\footnotesize},
    ylabel style={text width=4cm,align=center,font=\footnotesize},
    ylabel={Marginal effect on cognitive processing},
    legend style={
		at={(0.5,-1)},
		anchor=north,
  font=\footnotesize,
		legend columns=-1,
		/tikz/every even column/.append style={column sep=0.5cm}
	},
  ]
  \addplot+ [mark=*] table [x expr={\thisrow{x}-0.2}, y=y, y error=ey] {
    x y ey
1 2.951236647398 3.2333550618934
2 -8.88213223825815 5.25675198847529
3 -4.4382025760095 3.62793868470198
4 -4.79411664649971 4.09898236120032
5 -4.18493672923147 3.3863923258781
6 6.82513785669386 6.47777921841556
7 3.95266186265101 6.67912506323861
8 3.33410258633916 3.40384809391213
9 -0.493452382607021 3.71984845157033
10 4.43876184270656 4.56914996915033
11 0.635603223370301 4.85062439775103
12 -3.92209949769037 3.69472876978979
13 -0.975343639537532 4.0606098480258
14 5.68597108002593 4.5598121578347
15 -3.59823883858419 4.27013511979407
16 -0.92506865417821 3.6923742676095

  };
  \addplot+ [mark=*] table [x expr={\thisrow{x}+0.2}, y=y, y error=ey] {
    x y ey
1 3.66470410891778 4.91774039574824
2 0.791031197181532 3.22009236131875
3 -1.66238732182793 4.50872208690135
4 -1.48782177822592 3.88495842378226
5 1.74616370339638 4.92814635617449
6 -1.82430848768237 3.19700205456167
7 -1.19857613995951 3.13720426837057
8 -4.78819064764589 5.41947547244393
9 -1.66861100410241 4.02905886435274
10 8.26959746678816 3.51571128427588
11 0.708495074332554 3.62730696112097
12 6.8866504856762 4.30142158805362
13 4.06836466274348 4.09371421459093
14 0.529263503641044 3.56155493054954
15 1.83604215971009 4.03510332834873
16 -0.026463896948444 4.26748224072395

  };
  
\legend{Real News, Fake News}
\end{axis}
\end{tikzpicture}
}
\end{subfigure}
    \caption{Marginal effects of linguistic cues on cognitive processing (number of eye fixations). A larger number of eye fixations indicates that cognitive processing is more pronounced. Marginal effects are calculated based on the mixed-effects regression model in \Cref{eq:model_cog}. Control variables and subject-specific random effects are included. The error bars denote the 95\% confidence intervals.}
    \label{fig:marginal_cognitive}
\end{figure}
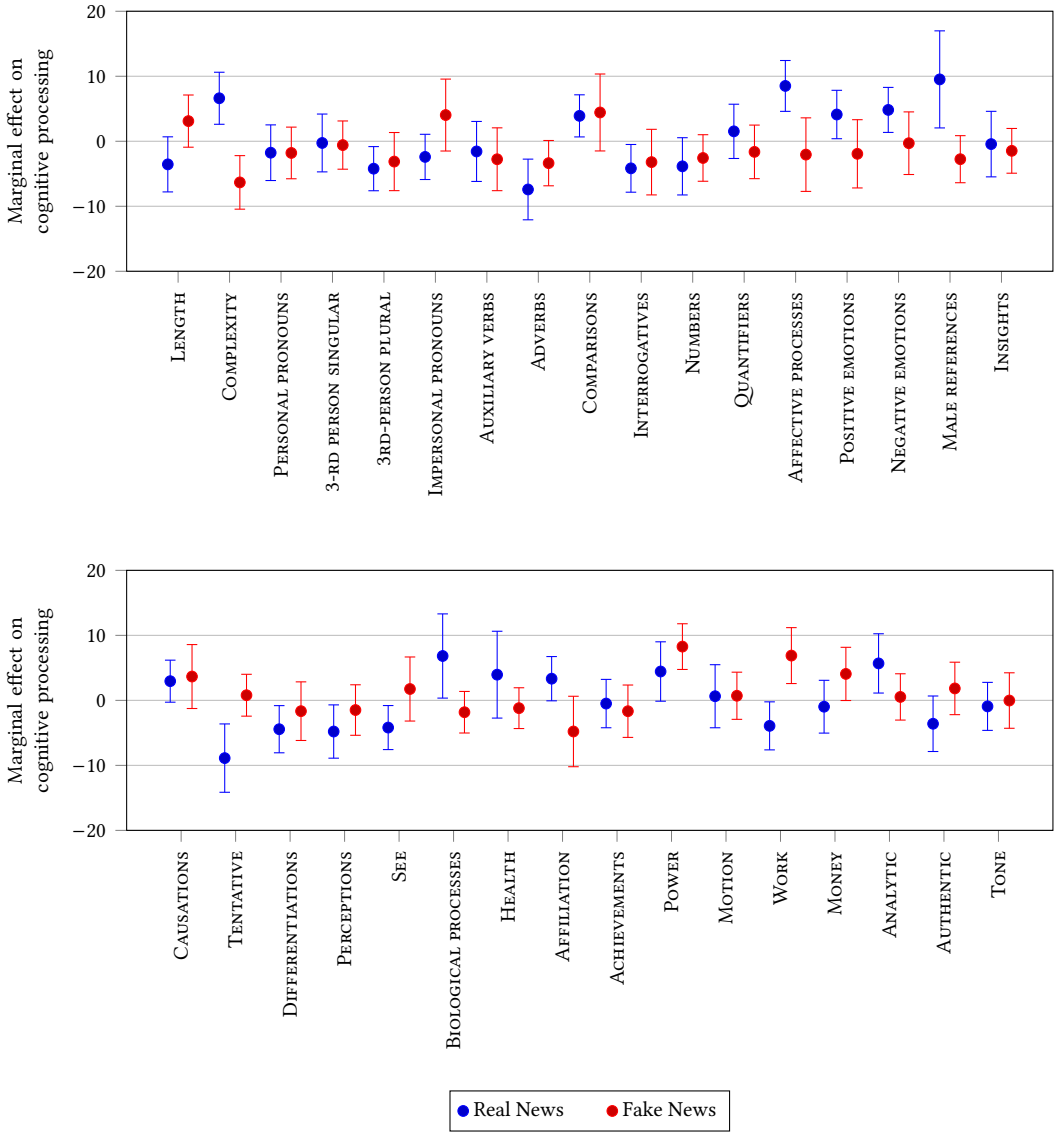

We first analyze how linguistic cues are linked to cognitive processing. 
Cognitive processing is measured as the number of eye fixations, \ie, a higher number of eye fixations indicates a greater cognitive effort. We use the mixed-effects model from \Cref{eq:model_cog} to calculate the marginal effects (ME) of linguistic cues on the number of eye fixations. Here, marginal effects measure the impact that one standard deviation change in one explanatory variable (\ie, a linguistic cue) has on the outcome variable (\ie, eye fixations) while all other variables are held constant. \Cref{fig:marginal_cognitive} shows the marginal effects\footnote{The numerical results of the marginal effects are provided in Appendix E of the Supplementary Materials.} of all linguistic cues for real news (blue) and fake news (red). The error bars denote the 95\% confidence intervals. The marginal effect of an explanatory variable is statistically significant if the 95\% confidence interval (given by its lower and upper boundaries denoted by the error bars) does not include zero. For instance, the marginal effects of personal pronouns on cognitive processing for real and fake news are not statistically significant, since both confidence intervals touch the zero line. However, if the lower and upper boundaries are both above or below zero, the marginal effect of the corresponding variable is significantly positive or negative, respectively.


Our results show that a total of 17 linguistic cues are statistically significantly linked to cognitive processing. For \textbf{real news}, we find that users engage more in cognitive processing if an article is written with greater \highlightcue{complexity}, more \highlightcue{comparisons}, more words linked to \highlightcue{affective processes}, \highlightcue{positive emotion words}, \highlightcue{negative emotion words}, \highlightcue{male references}, words linked to \highlightcue{biological processes}, and words reflecting \highlightcue{analytic thinking}. For instance, an increase of one standard deviation in the \highlightcue{complexity} of a real news article (SD=2.591) leads to 6.62 more eye fixations when reading and assessing the veracity of the article. In contrast, users engage less in cognitive processing of real news if an article is written with more \highlightcue{3rd-person plural pronouns}, \highlightcue{adverbs}, \highlightcue{interrogatives}, \highlightcue{tentative statements}, \highlightcue{differentiation}, words linked to \highlightcue{perceptual processes}, \highlightcue{seeing}, or \highlightcue{work-related} terms. For instance, if the usage of \highlightcue{adverbs} increases by one standard deviation (SD=1.634), then users perform on average 7.42 fewer eye fixations.

By contrast, for \textbf{fake news}, we find that users engage more in cognitive processing when an article is written with more \highlightcue{power} words or \highlightcue{work-related} terms. 
Conversely, cognitive processing of fake news is less pronounced if an article is written with lower \highlightcue{complexity}. In addition, we find weak evidence that cognitive processing of fake news is less pronounced when an article is written with a higher number of \highlightcue{adverbs}. If the usage of adverbs increases by one standard deviation in a fake news article (SD=1.634), then users perform 3.37 fewer eye fixations.



We make the following observations regarding the control variables in our model (see Appendix D of the Supplementary Materials for detailed results). First, users engage less in cognitive processing if the news article is labeled as fake according to official fact-checking. Specifically, users perform 7.967 fewer eye fixations when reading a fake news article as compared to a real news article. Second, we find that the reading time has a strong positive influence on cognitive processing, which is expected given that an increased reading time also allows for more eye fixations.

\subsubsection{Linguistic Cues and Affective Processing}

\begin{figure}[htp]
    \centering

    \begin{subfigure}[b]{\textwidth} {

    \begin{tikzpicture}
\pgfplotsset{
  error bars/.cd,
    x dir=none,
    y dir=both, y explicit,
}
\begin{axis}[
    width=\linewidth,
    height=5cm,
    only marks,
    ymajorgrids,
    tick pos=lower,
    tick align=outside,
    xmin=0, xmax=18,
    ymin=-5, ymax=5,
    xtick={1,2,3,4,5,6,7,8,9,10,11,12,13,14,15,16,17},
xticklabels={\highlightcue{Length},\highlightcue{Complexity},\highlightcue{Personal pronouns},\highlightcue{3-rd person singular},\highlightcue{3rd-person plural},\highlightcue{Impersonal pronouns},\highlightcue{Auxiliary verbs},\highlightcue{Adverbs},\highlightcue{Comparisons},\highlightcue{Interrogatives},\highlightcue{Numbers},\highlightcue{Quantifiers},\highlightcue{Affective processes},\highlightcue{Positive emotions},\highlightcue{Negative emotions},\highlightcue{Male references},\highlightcue{Insights}},
    y tick label style={font=\footnotesize},
    x tick label style={rotate=90, font=\footnotesize},
    xlabel style={font=\footnotesize},
    ylabel style={text width=4cm,align=center,font=\footnotesize},
    ylabel={Marginal effect on \strut affective processing},
    legend style={
		at={(0.5,-0.6)},
		anchor=north,
  font=\footnotesize,
		legend columns=-1,
		/tikz/every even column/.append style={column sep=0.5cm}
	},
  ]
  \addplot+ [mark=*] table [x expr={\thisrow{x}-0.2}, y=y, y error=ey] {
    x y ey
1 0.133732769630321 1.31870103367053
2 0.469406716871634 1.24538713144637
3 0.51081104817825 1.3300931191581
4 0.161423247550984 1.38367473367728
5 0.552721359689473 1.05743430026597
6 0.154426234630801 1.08067658520775
7 -0.719566068353474 1.43614765591391
8 -0.272613803323863 1.45791477660583
9 -0.496385433554653 1.00926078924594
10 -0.392254980855753 1.1433482536777
11 -0.509343975708913 1.37158198663955
12 -0.885957881386546 1.2982999295238
13 -0.170666461109535 1.22430340482525
14 0.41952508107997 1.15986106017458
15 -0.639826477645554 1.0794260599442
16 0.349502693056757 2.32880524560945
17 0.293906498017335 1.56305630350406

  };
  \addplot+ [mark=*] table [x expr={\thisrow{x}+0.2}, y=y, y error=ey] {
    x y ey
1 0.253217057995904 1.24941517820777
2 -0.557646513387818 1.28131108377919
3 1.37110584720155 1.23446254162341
4 -0.497929567571417 1.15411040699751
5 1.42963500674983 1.39238631968911
6 2.50305139887411 1.71877488294107
7 0.207147668403445 1.50418069113357
8 0.466549589456975 1.08859648917032
9 -1.24364296240389 1.84440117500437
10 1.2168079028337 1.57155255589901
11 -0.388035917396414 1.11534375708909
12 -0.234204915894475 1.28199307708613
13 0.586542354454293 1.77058084426468
14 0.53418521785139 1.63576849717366
15 0.147528562753622 1.50154381736427
16 -0.631997506218156 1.12908763453284
17 1.79819542606657 1.06739478633702

  };
  
\end{axis}
\end{tikzpicture}
}

\end{subfigure}
    \begin{subfigure}[b]{\textwidth} {

    \begin{tikzpicture}
\pgfplotsset{
  error bars/.cd,
    x dir=none,
    y dir=both, y explicit,
}
\begin{axis}[
    width=\linewidth,
    height=5cm,
    only marks,
    ymajorgrids,
    tick pos=lower,
    tick align=outside,
    xmin=0, xmax=17,
    ymin=-5, ymax=5,
    xtick={1,2,3,4,5,6,7,8,9,10,11,12,13,14,15,16},
xticklabels={\highlightcue{Causations},\highlightcue{Tentative},\highlightcue{Differentiations},\highlightcue{Perceptions},\highlightcue{See},\highlightcue{Biological processes},\highlightcue{Health},\highlightcue{Affiliation},\highlightcue{Achievements},\highlightcue{Power},\highlightcue{Motion},\highlightcue{Work},\highlightcue{Money},\highlightcue{Analytic},\highlightcue{Authentic},\highlightcue{Tone}},
    y tick label style={font=\footnotesize},
    x tick label style={rotate=90, font=\footnotesize},
    xlabel style={font=\footnotesize},
    ylabel style={text width=4cm,align=center,font=\footnotesize},
    ylabel={Marginal effect on \strut affective processing},
    legend style={
		at={(0.5,-1)},
		anchor=north,
          font=\footnotesize,
		legend columns=-1,
		/tikz/every even column/.append style={column sep=0.5cm}
	},
  ]
  \addplot+ [mark=*] table [x expr={\thisrow{x}-0.2}, y=y, y error=ey] {
    x y ey
1 -0.127132947221753 1.00773257707759
2 0.121108556486736 1.64076193388833
3 -0.729634878362766 1.12891306302232
4 -0.927045255936778 1.27673905341664
5 -0.961362519699435 1.05464805119779
6 -0.303570401278165 2.01640699658544
7 -0.804475523466419 2.07386297544284
8 0.242595484659249 1.06121605309957
9 0.733377196954768 1.1570410750151
10 0.757573794463427 1.43255882559165
11 0.503241707260439 1.50900907972932
12 1.05251395373962 1.15333446217
13 0.542299343082012 1.26472659680699
14 -0.0765844308998973 1.41277179001068
15 0.00813213380857876 1.32979787980035
16 0.565926554836724 1.14858747186625

  };
  \addplot+ [mark=*] table [x expr={\thisrow{x}+0.2}, y=y, y error=ey] {
    x y ey
1 0.233995482585809 1.53270073287742
2 0.599006571844446 1.00506859838146
3 1.48949750857694 1.40298041611752
4 0.451811863695284 1.2100994074206
5 0.316931450958129 1.53486354846209
6 1.02247449964576 0.995168126577316
7 1.38781335367826 0.974117218837142
8 -0.189444599374367 1.68965511320346
9 -0.266378744931337 1.25321598391688
10 0.0906622390448411 1.10227336869977
11 0.295971428848212 1.12844095841798
12 0.703205509161402 1.34271777900227
13 0.206103128656857 1.27503415126945
14 -2.46346734104398 1.10348016949304
15 -0.59387235919504 1.25661172323749
16 0.146006460736845 1.32748815091645

  };
  
\legend{Real News, Fake News}
\end{axis}
\end{tikzpicture}
}
\end{subfigure}
    \caption{Marginal effects of linguistic cues on affective processing (heart rate variability). Lower heart rate variability indicates that users experience an affective state of discomfort. Marginal effects are calculated based on the mixed-effects regression model in \Cref{eq:model_aff}. Control variables and subject-specific random effects are included. The error bars denote the 95\% confidence intervals.}
    \label{fig:marginal_affective}
\end{figure}
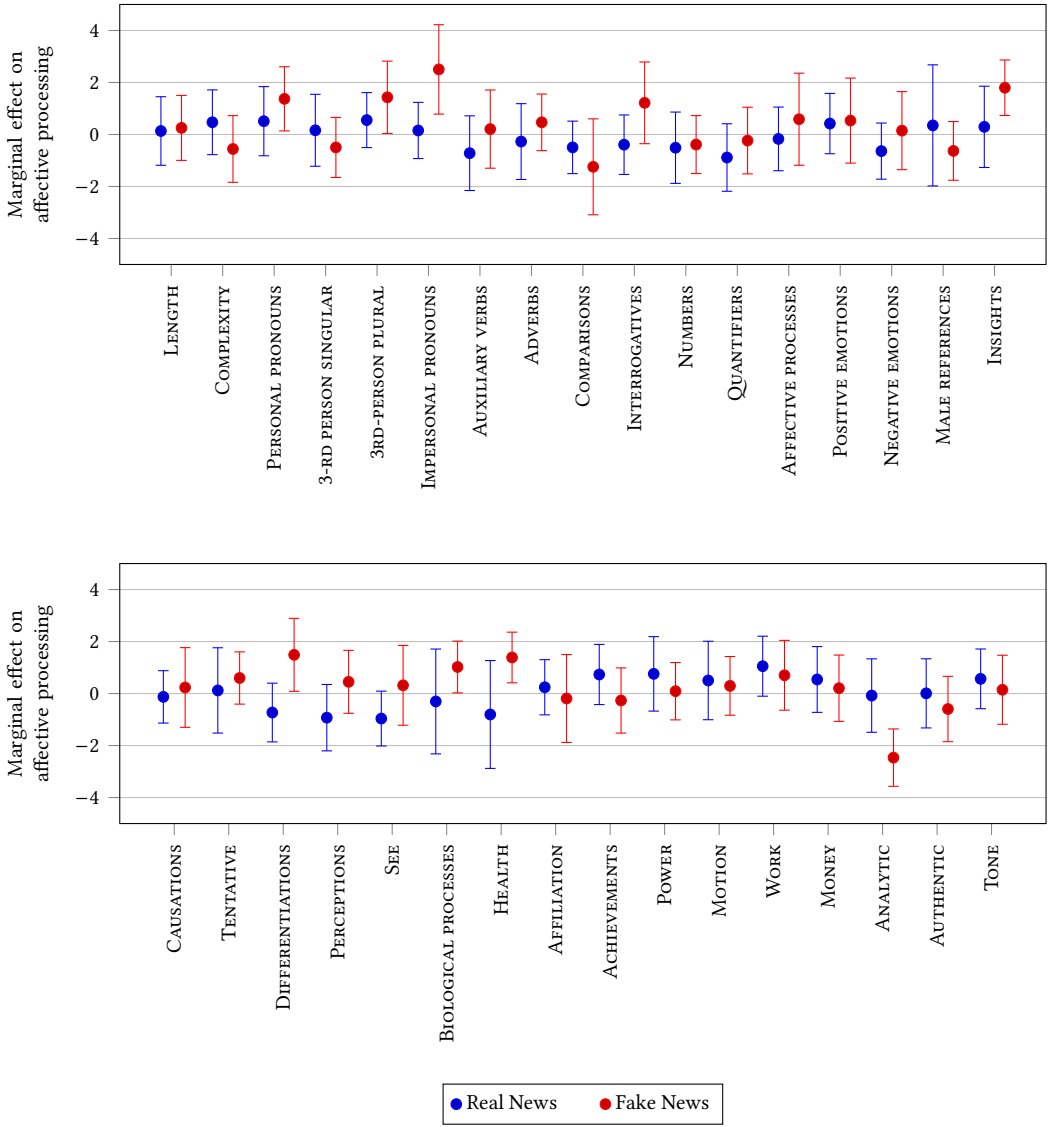

Next, we analyze the link between linguistic cues and affective processing. Affective processing is measured as heart rate variability in milliseconds (HRV). Lower HRV indicates that users experience an affective state of discomfort. We use the mixed-effects model from \Cref{eq:model_aff} to calculate the marginal effects (the numerical values are provided in Appendix E of the Supplementary Materials) of linguistic cues on the the HRV. \Cref{fig:marginal_affective} shows the marginal effects of linguistic cues for real and fake news along with the 95\% confidence intervals.

We find that a total of eight linguistic cues are statistically significantly linked to affective processing. Interestingly, we only observe significant estimates for affective processing of \textbf{fake news}, but none for \textbf{real news}. Specifically, users are less likely to experience an affective state of discomfort reflected by higher HRV if a fake news article is written with more \highlightcue{personal pronouns}, \highlightcue{3rd-person plural pronouns}, \highlightcue{impersonal pronouns}, \highlightcue{insights}, \highlightcue{differentiation}, words linked to \highlightcue{biological processes}, or \highlightcue{health}. In contrast, users are more likely to experience an affective state of discomfort when a fake news article is written with more terms linked to \highlightcue{analytic thinking}. Specifically, an increase of one standard deviation in the linguistic cues \highlightcue{analytic thinking} lowers users' heart rate variability by 2.64 milliseconds.



Regarding the control variables in our analysis (see Appendix D of the Supplementary Materials for detailed results), we find that affective processing does not depend on whether the article is labeled as fake according to fact-checking. The coefficient of the  the veracity dummy is not statistically significant. In addition, we observe that affective processing is less pronounced when the article has a higher sequence number (marginal effect: ME $=$ 0.290, $p < 0.001$), indicating potential habituation effects over time.

\subsection{Analysis of Socio-Demographics}


As an exploratory analysis, we consider the socio-demographic variables from the survey, including age, gender, cultural background (European or non-European), disposition to trust, news consumption, and political attitude. Recall that we accounted for subject heterogeneity by using subject-random effects as our focus was on within-subject effects. However, including socio-demographic variables allows us to analyze whether cognitive and affective processing depends on age, gender, or political attitude \citep[\eg,][]{Laserna.2014,Riedl.2010,Pennycook.2018}. We find that older,  European, and more liberal subjects engage more in cognitive processing. The same holds true for subjects with a higher disposition to trust and lower news consumption. On the other hand, younger, European, and more liberal subjects are more likely to experience an affective state of discomfort. The regression results using socio-demographics are provided in Appendix F of the Supplementary Materials.

\section{Discussion}

\subsection{Implications for Research}


We performed an experiment with neurophysiological measurements to understand if linguistic cues can help to explain how users process real and fake news. Identifying the driving factors of cognitive and affective processing allows for a deeper understanding of why users fall for fake news. While cognitive processing is considered crucial for correctly identifying fake news, affective processing involves automatic and unconscious evaluations. Hence, linguistic cues that trigger affective processing may make users may fall for fake news inadvertently and even without noticing. To the best of our knowledge, this is the first work studying the role of linguistic cues in fake news through a neurophysiological lens.

\newcommand{\SummarySymbol}[1]{\textbf{\Large{#1}}}

\newcommand{\SummarySymbolPlus}{\textbf{\color{darkgreen} \Large  $\boldsymbol{\oplus}$}}
\newcommand{\SummarySymbolMinus}{\textbf{\color{darkred} \Large $\boldsymbol{\ominus}$}}

\begin{table}[htp]
\caption{Summary of findings. Marginal effects of linguistic cues on cognitive and affective processing.\label{tbl:results_overview}}
\footnotesize
\begin{tabular}{>{\scshape}L{3cm}*{4}{C{2cm}}}
\toprule

\textnormal{\textbf{Linguistic cue}} & \multicolumn{2}{c}{
    \begin{tabular}{c}
         \textbf{Cognitive Processing}\\ \textbf{(eye fixations)}
    \end{tabular}} & \multicolumn{2}{c}{\begin{tabular}{c}
         \textbf{Affective Processing} \\ \textbf{(heart rate variability)}
    \end{tabular} } \\
\cmidrule(lr){2-3} \cmidrule(lr){4-5}
& \mc{\textbf{Real News}} & \mc{\textbf{Fake News}} & \mc{\textbf{Real News}} & \mc{\textbf{Fake News}} \\
\midrule 
Length & & \SummarySymbolPlus \\
Complexity & \SummarySymbolPlus & \SummarySymbolMinus \\

Personal pronouns & & & & \SummarySymbolPlus \\
3rd-person plural pronouns  & \SummarySymbolPlus  & & &  \SummarySymbolPlus \\
Impersonal pronouns &  & & &  \SummarySymbolPlus \\
Adverbs  & \SummarySymbolMinus  & & & \\
Comparisons &  \SummarySymbolPlus & & &  \\
Interrogatives &  \SummarySymbolMinus  & & & \\
Affective processes  & \SummarySymbolPlus  & & & \\
Positive emotion words & \SummarySymbolPlus  & & & \\
Negative emotion words & \SummarySymbolPlus  & & &\\
Male references  & \SummarySymbolPlus  & & &  \\
Insights &  & & &  \SummarySymbolPlus\\

Tentative &  \SummarySymbolMinus  & & &  \\
Differentiation & \SummarySymbolMinus  & & &  \SummarySymbolPlus \\
Perceptions & \SummarySymbolMinus \\
See & \SummarySymbolMinus\\
Biological processes & \SummarySymbolPlus &&& \SummarySymbolPlus \\
Health &&&& \SummarySymbolPlus \\
Power & \SummarySymbolPlus & \SummarySymbolPlus \\
Work & \SummarySymbolMinus & \SummarySymbolPlus\\
Analytic thinking & \SummarySymbolPlus  & & & \SummarySymbolMinus \\

\bottomrule
\multicolumn{5}{l}{Note that an affective state of discomfort is accompanied by \emph{lower} heart rate variability.} \\
\end{tabular}

\end{table}


Our findings are summarized in \Cref{tbl:results_overview}. We found that users engage more in cognitive processing of fake news articles when they have greater \highlightcue{length}. However, this finding does not apply to real news. One possible explanation is that fake news articles may provide more (fake) justifications to make a fake story appear more truthful, whereas real news articles may provide further factual details besides the actual information. The actual information in real news articles may also be described in a concise way, such that users require less cognitive effort to comprehend the content. The marginal effect of \highlightcue{complexity} has opposite signs for real and fake news. When the \highlightcue{complexity} of an article increases, users engage less in cognitive processing of real news, but more in the cognitive processing of fake news. In the literature on deception detection, documents with higher \highlightcue{complexity} are considered more truthful \citep[\eg,][]{Newman.2003,Zhou.2004}. Processing news articles with greater complexity also demands more cognitive resources. However, the subjects in our experiment did not engage more in cognitive processing when fake news articles are written with greater \highlightcue{complexity}. A lack of cognitive processing is an important factor of why users fall for fake news \citep{Moravec.2019,Pennycook.2020,Pennycook.2019}. Therefore, fake news articles written with greater \highlightcue{complexity} may be more successful in deceiving the public. Furthermore, we find that cognitive processing of fake news articles is more pronounced when an article is written with more \highlightcue{power} words or \highlightcue{work-related terms}. Accordingly, news articles with more \highlightcue{power} words or \highlightcue{work-related terms} may be less likely to deceive their readers.


Recent research has shown that, in addition to cognition, the processing of fake news is also governed by affect \citep{Lutz.2020,Lutz.2023}. The presence of affect can make users fall for fake news automatically -- and without even noticing which renders the role of affect highly concerning \citep{Lutz.2020,Lutz.2023}. However, it has remained unclear whether and how affective processes are triggered. Here, our study extends the literature by demonstrating that users experience an affective state of discomfort when fake news articles are written with more words reflecting \highlightcue{analytic thinking}. 
Words linked to \highlightcue{analytic thinking} can make an article appear more truthful as they clearly distinguish the content from what happened and what did not happen. There appear two possible explanations for  why users experience an affective state of discomfort if the content of a fake news article contradicts a user's worldview. Alternatively, users may experience discomfort in processing fake news when the article describes unpleasant events. 
We built upon neurophysiological measurements to study users' cognitive and affective processing, which is motivated by the fact that self-reported data often deviates from the underlying processes \citep{Walla.2018}. The reason is that humans cannot fully reflect on the affective processes occurring deep inside the brain \citep{vomBrocke.2020,Walla.2018}. In contrast, neurophysiological measurements can capture affective responses \emph{in situ}, i.e., directly at the moment when users process news. Thereby, we obtain complementary results that self-reports alone cannot capture \citep{Tams.2014}. To this end, our research question can \emph{only} be answered by using neurophysiological measurements.

\subsection{Implications for Practice}


Social media platforms need to be aware that real and fake news articles can differently trigger cognitive and affective processing in the user, subject to different linguistic cues of an article. Since linguistic cues can be calculated efficiently and reliably using existing tools from text mining, it is straightforward for social media platforms to gauge how different articles will be processed by the users. Based on that, platforms may prioritize content that elicits cognitive processing (as opposed to affective processing) in their ranking algorithms. Adapting the ranking and filtering algorithms of social media platforms in this way may present a cost-efficient and scalable countermeasure to reduce the likelihood that users fall for fake news. Notwithstanding these opportunities of social media platforms, regulators could also mandate that social media platforms account for cognitive-processing-inducing cues. 


In particular, social media platforms, users, and politicians need to be cautious of fake news written with greater \highlightcue{complexity} and more words linked to \highlightcue{analytic thinking}. Fake news articles with greater \highlightcue{complexity} may appear truthful, while requiring more cognitive effort to be processed. Such fake news articles may thus increase users' propensity to fall for fake news. In addition, fake news articles can be subject to automatic and unconscious evaluations, if they are written with words linked to \highlightcue{analytic thinking}. This means that users may fall for such fake news articles even without realizing it. Social media platforms should therefore be alerted when news articles of unclear veracity are characterized by greater \highlightcue{complexity} and more \highlightcue{analytic words}. Accordingly, they should consider designing their platforms in a way that prevents users from being exposed to these articles.

Many commercial large language models (LLMs) use a professional language characterized by analytic cues, and it is thus likely that our findings also apply to why users are prone to fall for hallucinations of LLMs. Furthermore, LLMs could even be employed by malicious actors to compose fake news \cite{NHB} in a way that the above-mentioned psychological dynamics are strategically targeted to elicit certain beliefs, which poses a serious concern that could be addressed by better media literacy training and better algorithmic curation on social media platforms (as discussed above).

\subsection{Limitations and Future Research}



Our study suffers from the general limitations of laboratory studies. Our experiment was conducted in the context of a neutral laboratory that differs from how subjects are used to reading online news articles in their everyday life. In addition, our subjects were undergraduate students, which may not be representative of the whole society \citep{Koriat.1980}. Furthermore, the reward for positive identification of real and fake news could have induced a utilitarian mindset, while users on social media can generally be assumed to be in a hedonic mindset \citep{Moravec.2019}. Consequently, the presence of affective information processing could be even more pronounced in real-world scenarios. To address this issue, conducting field experiments utilizing wearable devices for measuring HRV presents a potential avenue for future research. However, the reliability and accuracy of measurements obtained from wearable devices remain subjects of debate \citep{Stangl.2022}. Moreover, the adoption of such an approach raises privacy concerns, as it would necessitate the recording of social media interactions. As such, it is common practice for related studies on fake news to use lab settings \citep{Moravec.2019,Moravec.2022}.

\subsection{Conclusion}

Understanding how real and fake news is processed is important for curbing the spread of fake news. Cognitive processing is considered a necessary requirement to prevent users from falling for fake news, while affective processing occurs automatically and unconsciously, thereby increasing the likelihood of less elaborated decisions. Linguistic cues are an important tool in deception detection and they can easily be measured using established text analysis software. However, little is known about how linguistic cues influence users' cognitive and affective processing. We performed a lab experiment with eye-tracking and heart rate measurements to explain cognitive and affective processing of real and fake news based on linguistic cues. Our results suggest that cognitive processing of fake news is less pronounced if an article is written with lower \highlightcue{complexity} or more \highlightcue{adverbs}. In addition, we find that affective processing of fake news is more pronounced when an article is written with greater usage of terms reflecting \highlightcue{analytic thinking}. Social media platforms aiming to curb the spread of fake news should hence be cautious when news articles are written with low \highlightcue{complexity}, a high number of \highlightcue{adverbs}, or high number of terms associated with \highlightcue{analytic thinking}. To the best of our knowledge, this is the first study employing neurophysiological measurements to explain human information processing according to linguistic cues.

\section*{Acknowledgement}

We appreciate funding from the German Academic Exchange Service (grant number 57317901).

\bibliographystyle{ACM-Reference-Format}
\bibliography{references}

\end{document}